%% file: main.tex
\documentclass[runningheads]{llncs}

\usepackage[T1]{fontenc}
\usepackage{graphicx}
\usepackage{enumitem}
\usepackage{booktabs}
\usepackage{pgfgantt}
\usepackage{appendix}
\usepackage{enumitem}
\usepackage{algorithm}
\usepackage{algpseudocode}
\usepackage{array}
\usepackage{tikz}
\usetikzlibrary{shapes.geometric, arrows, fit}
\usepackage{pgfplots}
\usepackage{url}
\usepackage{amsmath}
\usepackage{orcidlink}
\usepackage{comment}

\begin{document}

\title{Evaluating LLM-Based Process Explanations under Progressive Behavioral-Input Reduction}
\titlerunning{LLM-Based Process Explanations under Behavioral-Input Reduction}

\author{P. van Oerle\,\orcidlink{0009-0004-8084-8622}
\and R.H. Bemthuis\,\orcidlink{0000-0003-2791-6070}
\and F.A. Bukhsh\,\orcidlink{0000-0001-5978-2754}}
\authorrunning{P. van Oerle et al.}

\institute{University of Twente, Enschede, The Netherlands \\
\email{p.vanoerle@student.utwente.nl} \\
\email{\{r.h.bemthuis,f.a.bukhsh\}@utwente.nl}}

\maketitle

\begin{abstract}
Large Language Models (LLMs) are increasingly used to generate textual explanations of process models discovered from event logs. Producing explanations from large behavioral abstractions (e.g., directly-follows graphs or Petri nets) can be computationally expensive. This paper reports an exploratory evaluation of explanation quality under progressive behavioral-input reduction, where models are discovered from progressively smaller prefixes of a fixed log. Our pipeline (i) discovers models at multiple input sizes, (ii) prompts an LLM to generate explanations, and (iii) uses a second LLM to assess completeness, bottleneck identification, and suggested improvements. On synthetic logs, explanation quality is largely preserved under moderate reduction, indicating a practical cost-quality trade-off. The study is exploratory, as the scores are LLM-based (comparative signals rather than ground truth) and the data are synthetic. The results suggest a path toward more computationally efficient, LLM-assisted process analysis in resource-constrained settings.

\end{abstract}

\section{Introduction}
\label{section:introduction}
Process mining extracts actionable insights from event logs by discovering process models~\cite{book-aalst-2007,emamjome}. Over the last two decades, tools for discovery and conformance checking have matured substantially~\cite{viner2021processminingsoftwarecomparison,pminthelarge}, with applications in logistics, healthcare, and finance~\cite{bpmindustrialapplication}. Recent advances in large language models (LLMs) enable automatic generation of such explanations~\cite{surveyonevaluationofllm,wang2024history}. However, producing explanations from behavioral abstractions derived from large logs, such as directly-follows graphs (DFGs) or Petri nets, can be computationally intensive partly due to the size and structural complexity of the input given to the model. While each event may contribute unique behavior that is, in principle, worth capturing, it remains an open question how much input is required to obtain high-quality explanations. Here, we use behavioral abstraction to denote formal structures (e.g., control-flow graphs) derived from event logs and supplied to an LLM as the basis for textual descriptions.

Despite promising results with models such as GPT-4 and Llama for insights and next-step prediction~\cite{berti2024evaluatingllminpm,jessen2023chitchatdeeptalkprompt}, leveraging LLMs for explanations remains challenging in data-intensive settings. Large logs require substantial computation~\cite{pminthelarge}, motivating behaviorally efficient approaches that reduce the size of the abstraction while retaining explanation fidelity. Prior work examines LLM capabilities in process mining but rarely studies the effect of reducing behavioral information~\cite{howmuchdataisenough}. Recent works highlight both promise and limitations across semantic awareness~\cite{rebmann2024evaluating}, anomaly detection, and prediction~\cite{berti2023leveraginglargelanguagemodels,grohs2023largelanguagemodelsaccomplish}, yet to our knowledge do not quantify the trade-off between input complexity and explanation quality. 

This paper reports an exploratory evaluation of explanation quality under progressive behavioral-input reduction, where process abstractions are constructed from progressively fewer events. Studying reduction is relevant for: (i) data efficiency (e.g., acceptable explanations may be achievable from smaller abstractions); (ii) non-linear payoff (e.g., quality gains may plateau beyond mid-range input sizes); (iii) robustness (e.g., omitting rare variants or later events probes resilience under constraints); and (iv) practical thresholds (e.g., organizations need guidance on input sizes that balance quality with usability). 

We fix the log and vary the fraction of events used to construct the model, thereby potentially lowering the complexity of the structures provided to the LLM. Concretely, we propose a pipeline that (i) discovers process models from progressively smaller subsets of synthetic logs, (ii) prompts an LLM to produce textual explanations, and (iii) uses a second LLM to evaluate those explanations along three dimensions: completeness, bottleneck identification, and suggested improvements. 

Our contributions are: (1) a pipeline for evaluating the effect of behavioral-input reduction on LLM-based process explanations; (2) an exploratory quantification of explanation quality along three dimensions (completeness, bottlenecks, improvements) under progressive behavioral reduction; and (3) empirical evidence on synthetic logs that identifies a region of favorable cost--quality trade-offs, informing resource-constrained and real-time process intelligence systems.

The remainder of this paper is structured as follows. Section~\ref{section:relatedwork} reviews related work. Section~\ref{section:pipeline} presents our research pipeline for data reduction. Section~\ref{section:experiment} details the experimental setup. Section~\ref{section:results} reports and briefly discusses the results. Lastly, Sect.~\ref{section:threats} discusses threats to validity, and Sect.~\ref{section:conclusion} concludes. 

\section{Related Work}
\label{section:relatedwork}
Recent work at the intersection of LLMs and process mining explores tasks such as semantically aware process discovery~\cite{rebmann2024evaluating} and next-step prediction~\cite{berti2024evaluatingllminpm,jessen2023chitchatdeeptalkprompt}. LLM-based methods can exploit contextual language cues that reveal subtle operational details. However, existing works~\cite{berti2024pmllmbenchmarkevaluatinglargelanguage,berti2023leveraginglargelanguagemodels} largely emphasize accuracy or descriptive clarity and do not explicitly study data efficiency or the robustness of explanations. 

Beyond algorithmic enhancements, event log pre‑processing, covering tasks such as filtering, noise removal and sampling, has long been pivotal for scalable analysis~\cite{eventlogpreprocessing}. Yet little work combines such preprocessing with LLM-driven interpretation, despite growing volumes of heterogeneous logs where computation can be a bottleneck. We take a step in this direction by asking whether high quality explanations can be obtained from reduced behavioral abstractions. Online/streaming process mining addresses ingest-time scalability (e.g., single-pass or windowed processing)~\cite{pminthelarge}, while our focus is on how much behavioral abstraction is sufficient for high quality LLM explanations once a log (or its prefix) is available. 

In parallel, LLMs have been applied to process improvement. Lashkevich et al.~\cite{lashkevich2024llm} propose a prompting method in which an LLM analyzes waiting time causes and recommends redesign options. In a user study, an enhanced prompt with a structured catalog of redesign patterns yields more relevant and actionable suggestions than a zero-shot baseline while maintaining diversity and novelty. Kubrak et al.~\cite{kubrak2024explanatory} study LLM-generated explanations for process monitoring. Using a design science approach and an explainability question bank, they find dialogue-style explanations aid understanding but often fail to address the “why” and to support trust calibration. These studies indicate value for improvement and explainability, but assume full access to behavioral information and do not examine reduced abstractions. 

In contrast, we study the data efficiency of LLM-generated explanations. By systematically reducing the fraction of events used to build the behavioral model and quantifying the effect on explanation quality, we extend LLM-assisted process mining to reduction regimes and surface trade-offs between model complexity and explanation quality, which may be particularly relevant for large-scale or resource-constrained process intelligence. 

\section{Research Pipeline}
\label{section:pipeline}
This section presents a four-stage pipeline to test whether LLMs can be applied more data-efficiently in process mining. Figure~\ref{fig:pipeline} summarizes the flow. We vary the behavioral input used for model discovery and measure the impact on explanation quality. 

Our goal is not to reduce the number of logs, but to progressively reduce the behavioral footprint derived from a given log by selecting only a fraction of its events prior to discovery. 

\input{additions/processdiagram}

\subsection{Event Logs and Sub-log Generation}
\label{section:sublogs}
Let $L$ denote the full event log, represented as a time-ordered sequence of events $L=\langle e_1, e_2, \ldots, e_n\rangle$ with $|L|=n$. For $k\in\{1,\ldots,n\}$, a sub-log $S_k$ is the prefix containing the first $k$ events:

\begin{equation}
    S_k \;=\; \langle e_1, e_2, \ldots, e_k\rangle,\quad k\in\{1,\ldots,n\}.
\end{equation}

This prefix sampling is simple and reproducible, but it may exclude infrequent variants or late-appearing behavior; we discuss implications and alternatives in Section~\ref{section:threats}. Other strategies (e.g., trace-based subsampling, variant-aware or stratified selection) are compatible with the pipeline and are left for future work. 

\subsection{Process Discovery}
Process discovery maps an event log to a process model. Applying a discovery algorithm to $S_k$ yields: 

\begin{equation}
    M_k \;=\; \operatorname{Discover}(S_k),
\end{equation}

and $M=\operatorname{Discover}(L)$ for the full log. The pipeline is agnostic to the specific discovery method.

\subsection{LLM$_1$: Explanation Generation}
For each discovered model $M_k$, the first LLM (LLM$_1$) generates a textual explanation: 

\begin{equation}
    E_k \;=\; \text{LLM}_1(M_k).
\end{equation}

Each $E_k$ is prompted to (i) describe the process, (ii) identify bottlenecks, and (iii) suggest improvements.

\subsection{LLM$_2$: Scoring Against the Full Model}
A second LLM (LLM$_2$) evaluates $E_k$ with the full model $M$ as reference: 

\begin{equation}
    \mathrm{Sc}_k \;=\; \text{LLM}_2(E_k, M).
\end{equation}

Concretely, LLM$_2$ assigns three scores in $[1,10]$ (related to completeness, bottleneck identification, and improvement suggestions), then returns their average as the overall score. This setup indicates how well $E_k$ captures salient behavior relative to $M$. 

This pipeline enables a controlled study of how decreasing behavioral input (in $M_k$) affects the quality of generated explanations. While LLM-as-judge raises concerns about determinism and validity, it is practical for multi-faceted, semantically rich criteria that can be challenging to formalize. 

\section{Experimental Configuration}
\label{section:experiment}
This section details the experimental setup: the case study, data preprocessing, and LLM configuration. 

\subsection{Case Study Introduction}
We use the synthetic dataset from~\cite{bemthuis2021data}, which contains event logs generated via discrete event simulation of a job-shop scheduling problem. Each experiment provides raw results and a corresponding \texttt{.xes} log with roughly 1{,}000–1{,}500 traces (about 250{,}000–350{,}000 events). We convert logs to process models using standard tools; in our experiments we employ the Inductive Miner~\cite{leemans2014process} as implemented in PM4Py~\cite{berti2019processminingpythonpm4py}, due to its robustness on noisy, large-scale logs.

\subsection{Data Preprocessing}
\label{section:preprocessing}
All \texttt{.xes} logs are processed with PM4Py~\cite{berti2019processminingpythonpm4py}. From each log, we derive partial models $M_k$ for

\[
k \in \{10, 20, 50, 100, 1000, 10000, 100000\},
\]

as well as the full model $M$. These $k$ values span a wide range of behavioral inputs. Note that $k$ counts events used for discovery, not traces or logs. Table~\ref{tab:tracesandlogs} summarizes the datasets.

\begin{table}[ht]
\centering
\caption{Number of traces and events per experiment.}\label{tab:tracesandlogs}
\begin{tabular}{|c|c|c|}
\hline
Experiment & Number of traces & Number of events \\
\hline
411 & 1{,}296 & 314{,}160 \\
412 & 1{,}013 & 244{,}384 \\
413 & 1{,}308 & 317{,}072 \\
421 & 1{,}422 & 345{,}520 \\
422 & 1{,}223 & 296{,}184 \\
\hline
\end{tabular}
\end{table}

\subsection{LLM Configuration}
All experiments were conducted in Python. We accessed Meta's Llama 3.3-70B~\cite{grattafiori2024llama3herdmodels} via standard APIs. Llama was chosen for its balance of quality and cost. 

For each experiment in Table~\ref{tab:tracesandlogs} and for each $M_k$, we generate five explanations with independent runs, then score each explanation using a separate LLM instance on a 1–10 scale; we report the average per $k$. The generation prompt asks for: (i) a concise process description, (ii) bottleneck identification, and (iii) actionable improvement suggestions. An explanation is satisfactory if it covers the end-to-end flow, highlights inefficiencies, and proposes concrete actions. 

Table~\ref{tab:scoring} lists the scoring rubric. The evaluation prompt assigns three scores (completeness, bottlenecks, improvements) and returns their mean as the overall score. Code and results are publicly available\footnote{\url{https://github.com/oerlep/data-reduction-llm-in-pm/}}. Notice that the rubric serves as guidance for the evaluator LLM; we analyze scores comparatively across $k$ rather than as absolute ground truth.

\begin{table}[ht]
\centering
\caption{Scoring criteria for each metric.}\label{tab:scoring}
\resizebox{0.80\textwidth}{!}{%
\begin{tabular}{|c|c|}
\hline
Score & Criteria\\
\hline
1-3 &  Major elements are not mentioned or incorrect information is given. \\
4-6 &  All main elements are mentioned, but lack detail or are incomplete. \\
7-9 & All important elements are mentioned with minor omissions. \\
10 & All aspects are accurately described without omissions. \\
\hline
\end{tabular}
}
\end{table}

\section{Experimental Results}
\label{section:results}
This section reports the key findings. We first describe the explanations of the LLM's responses, then examine how explanation quality varies with the size of the input data. 

\subsection{Explanations of Discovered Models}
Generated explanations include: (i) a concise process overview (start/end points and major activities); (ii) control-flow structures (sequences, parallelism, loops); (iii) structural bottleneck candidates (e.g., self-loops, high-wait transitions); and (iv) actionable improvement suggestions. 

High-scoring outputs reference all major activities and propose concrete improvement strategies. Low-scoring outputs typically omit key steps or provide vague recommendations (e.g., “investigating causes”) without specific actions or data references.

\subsection{LLM-based Judgments}
The evaluator LLM assigns three scores (1–10) per explanation: completeness, bottleneck identification, and improvement suggestions. Explanations should cover main activities, transitions, and loops; identify salient performance issues; and propose data-informed changes. Missing or unclear elements reduce the score. 

LLM-based scoring is efficient but introduces variability and subjectivity. To mitigate this, we generate five explanations per $k$ and report the average. Section~\ref{section:threats} discusses complementary evaluation strategies.

\subsection{Effect of Behavioral Input Size on Quality}
Figure~\ref{fig:results} shows the average score as a function of $k$. Table~\ref{tab:results1} provides per-experiment details. Here, $k$ is the number of events used to discover the behavioral model, not the number of traces or logs. We observe: 

\begin{itemize}
    \item \textbf{Small $k$ (10–20):} low average scores ($<\!5.0$), reflecting incomplete or imprecise explanations. 
    \item \textbf{Mid-range $k$ (100–1{,}000):} substantial quality gains, with typical averages around 6.5–7.0. 
    \item \textbf{Large $k$ (10{,}000–100{,}000):} further improvements and higher consistency, stabilizing near $\approx 8.0$. 
\end{itemize}

\input{additions/graph}

\input{additions/tables}

Across experiments, the curve shows a practical knee between $k{=}100$ and $k{=}1{,}000$: average scores rise from $\approx6.5$ to $\approx6.9$. From $1{,}000$ to $10{,}000$, gains are modest (to $\approx7.0$), suggesting diminishing returns in that range under our setup. By contrast, moving to $k{=}100{,}000$ yields a clearer uplift (to $\approx7.92$) and visibly narrower error bars, indicating higher consistency. The full logs provide only a marginal additional increase ($\approx7.95$ on average), likely within the variability observed. Thus, $k\approx 1{,}000$ offers a stable mid-range when latency or budget constraints apply, while $k\approx 100{,}000$ can be justified in higher-stakes settings that benefit from the extra point of quality. Per-experiment trends are monotonic on average, though individual logs show slight non-monotonicity around $k{=}10{,}000$, illustrating the value of reporting variability. These findings remain contingent on synthetic job-shop data and may vary with real-world noise and variant structure. 

\subsection{Discussion}
\label{subsection:discussion}
The results suggest that meaningful explanations can be derived from significantly reduced behavioral input, paving the way for more scalable and resource-efficient LLM-assisted process intelligence. Explanation quality improves non-linearly with input size: from very small sub-logs to mid-range values of $k$, we observe clear gains; between $k = 1{,}000$ and $k = 10{,}000$, improvements are modest under our setup; at $k = 100{,}000$, we see a further uplift along with reduced variance, while full logs yield only marginal additional improvement. A plausible explanation can be coverage: very small sub-logs under-sample rare variants and loop structures, resulting in under-specified models. Larger values of $k$ capture more control-flow detail, enabling richer explanations. The precise “sweet spot” may depend on the task and domain; for example, sensitivity to rare events or tolerance for cost and latency. 

We used Inductive Miner with default settings and serialized the discovered Petri nets for LLM input. Inductive Miner's block-structured nets are sound and may over-approximate behavior, which can improve readability but alter the abstraction exposed to the LLM. We expect notation and algorithm choices (e.g., DFGs or BPMN; Heuristics/Alpha Miner; intentionally unsound nets) to affect explanation quality by changing label granularity and control-flow clarity. A systematic analysis over discovery methods and model forms is an important next step.

\section{Threats to Validity}
\label{section:threats}
This study has several limitations that may affect the generalizability, reproducibility, and interpretability of the results. 

Using an LLM to grade another LLM introduces subjectivity, prompt sensitivity, and potential bias propagation. We averaged five generations per $k$, yet residual variability remains, and evaluator preferences (e.g., phrasing/style) may correlate with the generator's output. Even for full logs, no perfect (10) scores were produced, suggesting missing context or model limitations. We therefore treat scores as comparative signals, not ground truth. Future work should triangulate with (i) human ratings by process analysts, (ii) cross-model judging (e.g., different evaluator LLMs), and (iii) task-specific objective proxies (e.g., activity/edge coverage against $M_k$, penalties for hallucinated activities, alignment-based fitness of described paths). We plan a small expert study (think-aloud and rubric) to calibrate LLM scores and assess inter-rater agreement. 

Prefix sampling (first $k$ events) simplifies incremental comparisons but may under-sample late or rare variants, potentially inflating apparent gains at larger $k$. Alternative strategies (e.g., variant-aware, stratified, or trace-level sampling) could alter the curve. Prompt wording, decoding parameters, and seed control affect both generations and judgments. We report averages but acknowledge sensitivity to these settings. Results may also depend on discovery and notation choices. 

With five generations per point, uncertainty persists. We did not report significance tests or confidence intervals. Replications could include dispersion (e.g., bootstrap CIs), effect sizes, and corrections for multiple comparisons across $k$ and datasets. 

Findings are based on synthetic job-shop logs, aiding control and reproducibility but not fully reflecting real-world noise, heterogeneity, or semantic ambiguity. Validation on organizational logs across domains is needed to assess robustness and to recalibrate the cost–quality frontier under realistic conditions. 

Evolving LLMs (with e.g., model updates, API changes) can affect replication. We release code, prompts, and outputs, but exact reproduction may vary with model versions. Recording model identifiers, decoding parameters, and system prompts, and including stable open-weight baselines alongside proprietary models, would further improve reproducibility.

\section{Conclusion}
\label{section:conclusion}
We examined whether useful LLM-based process explanations can be generated from reduced behavioral input. Using synthetic event logs, we discovered models from progressively smaller prefixes and evaluated the resulting explanations with a second LLM. Our results indicate that explanation quality improves with larger inputs but does not scale linearly, with diminishing returns once a sufficient behavioral footprint is captured. This suggests the existence of practical cost–quality trade-offs: smaller inputs may already yield adequate explanations for many scenarios, while larger inputs can provide additional consistency in settings where this is critical.

While our pipeline offers a preliminary artifact for data-efficient LLM applications in process mining, several caveats remain. This study is exploratory. Scores are produced by an LLM acting as judge and should be interpreted as comparative signals rather than ground truth. Furthermore, results are based on synthetic job-shop data and may not generalize to noisier, heterogeneous logs. The full model is used here only as an evaluation reference, not as a deployment requirement. 

Future work includes validation on real-world logs; analyzing over discovery algorithms and model notations; representative sampling (e.g., variant-aware or stratified) to preserve coverage at smaller $k$; and human and cross-LLM evaluations alongside objective proxies (e.g., activity/edge coverage, hallucination penalties), with uncertainty reporting and runtime/cost profiles. The reduction-first pipeline can also be applied to related tasks (e.g., improvement recommendation, prescriptive monitoring explanations) to study data efficiency more broadly. 

\subsubsection*{Disclosure of Interests}
The authors have no competing interests to declare that are relevant to the content of this article. 

\bibliographystyle{splncs04}
\bibliography{references}

\end{document}

%% file: additions/processdiagram.tex
\tikzstyle{process} = [rectangle, minimum width=2cm, minimum height=1cm, text centered, draw=black]
\tikzstyle{wideprocess} = [rectangle, minimum height=1cm, text centered, draw=black]
\tikzstyle{filled} = [rectangle, minimum height=1cm, text centered, draw=black, fill=gray!20]
\tikzstyle{arrow} = [thick,->]

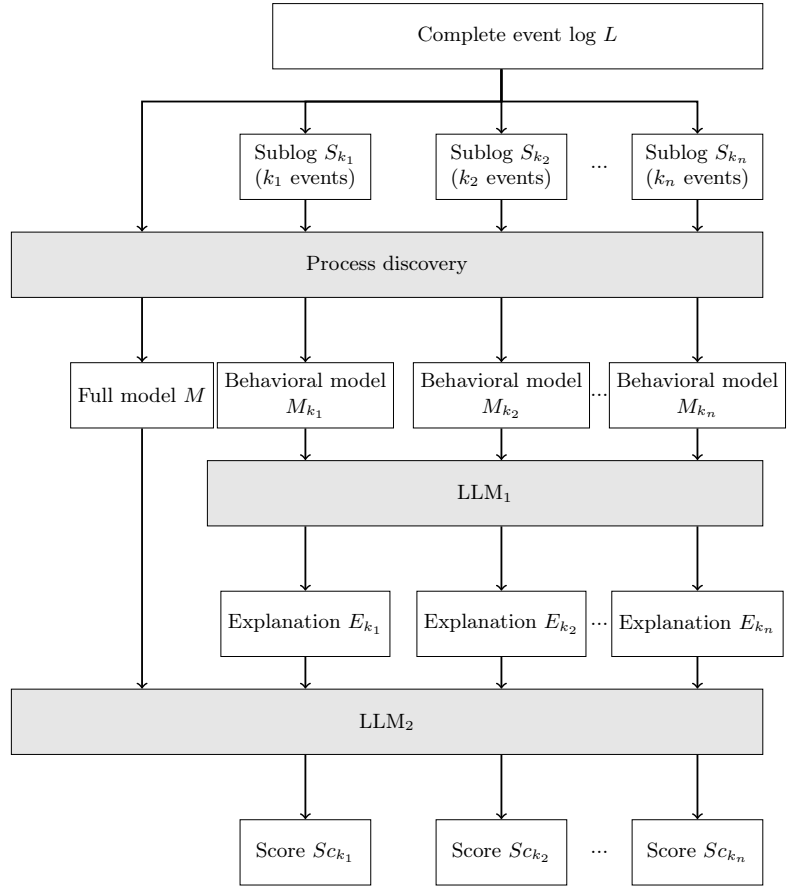
\begin{figure}[ht]
    \centering
    \resizebox{0.86\linewidth}{!}{
        \begin{tikzpicture}[node distance=1.5cm and 2cm, align=center]

            \node (fulllogs) [process, minimum width=7.5cm] {Complete event log $L$};
            
            \node (n2) [process, below of=fulllogs, yshift=-0.5cm, xshift=-0.25cm] {Sublog $S_{k_2}$\\($k_2$ events)};
            \node (sublogn1) [process, left of=n2, xshift=-1.5cm] {Sublog $S_{k_1}$\\($k_1$ events)};
            \node (dots1) [right of=n2, xshift=0cm] {...};
            \node (nn) [process, right of=dots1, xshift=0cm] {Sublog $S_{k_n}$\\($k_n$ events)};
            
            \node (processdiscovery) [filled, below of=sublogn1, minimum width=11.5cm, xshift=1.25cm] {Process discovery};
            
            \node (n2b) [process, below of=n2, yshift=-2cm] {Behavioral model\\$M_{k_2}$};
            \node (fullmodel) [process, left of=n2b, xshift=-4cm] {Full model $M$};
            \node (submodeln1) [process, left of=n2b, xshift=-1.5cm] {Behavioral model\\$M_{k_1}$};
            \node (dots2) [right of=n2b, xshift=0cm] {...};
            \node (nnb) [process, right of=dots2, xshift=0cm] {Behavioral model\\$M_{k_n}$};
            
            \node (llm1) [filled, below of=submodeln1, minimum width=8.5cm, xshift=2.75cm] {LLM$_1$};
            
            \node (n2c) [process, below of=n2b, yshift=-2cm] {Explanation $E_{k_2}$};
            \node (expn1) [process, left of=n2c, xshift=-1.5cm] {Explanation $E_{k_1}$};
            \node (dots3) [right of=n2c, xshift=0cm] {...};
            \node (nnc) [process, right of=dots3, xshift=0cm] {Explanation $E_{k_n}$};
            
            \node (llm2) [filled, below of=expn1, minimum width=11.5cm, xshift=1.25cm] {LLM$_2$};
            
            \node (n2d) [process, below of=n2c, yshift=-2cm] {Score $Sc_{k_2}$};
            \node (scoren1) [process, left of=n2d, xshift=-1.5cm] {Score $Sc_{k_1}$};
            \node (dots4) [right of=n2d, xshift=0cm] {...};
            \node (nnd) [process, right of=dots4, xshift=0cm] {Score $Sc_{k_n}$};
            
            \draw [arrow] (n2.north |- fulllogs.south) -- (-0.25, -1) -| (sublogn1);
            \draw [arrow] (n2.north |- fulllogs.south) -- (-0.25, -1) -| (n2);
            \draw [arrow] (n2.north |- fulllogs.south) -- (-0.25, -1) -| (nn);
            \draw [arrow] (n2.north |- fulllogs.south) -- (-0.25, -1) -| (fullmodel.north |- processdiscovery.north);
            
            \draw [arrow] (sublogn1.south) -- (sublogn1.south |- processdiscovery.north);
            \draw [arrow] (n2.south) -- (n2.south |- processdiscovery.north);
            \draw [arrow] (nn.south) -- (nn.south |- processdiscovery.north);
            
            \draw [arrow] (fullmodel.north |- processdiscovery.south) -- (fullmodel.north);
            \draw [arrow] (submodeln1.north |- processdiscovery.south) -- (submodeln1.north);
            \draw [arrow] (n2b.north |- processdiscovery.south) -- (n2b.north);
            \draw [arrow] (nnb.north |- processdiscovery.south) -- (nnb.north);
            
            \draw [arrow] (fullmodel.south) -- (fullmodel.south |- llm2.north);
            \draw [arrow] (submodeln1.south) -- (submodeln1.south |- llm1.north);
            \draw [arrow] (n2b.south) -- (n2b.south |- llm1.north);
            \draw [arrow] (nnb.south) -- (nnb.south |- llm1.north);
            
            \draw [arrow] (expn1.north |- llm1.south) -- (expn1.north);
            \draw [arrow] (n2c.north |- llm1.south) -- (n2c.north);
            \draw [arrow] (nnc.north |- llm1.south) -- (nnc.north);
            
            \draw [arrow] (expn1.south) -- (expn1.south |- llm2.north);
            \draw [arrow] (n2c.south) -- (n2c.south |- llm2.north);
            \draw [arrow] (nnc.south) -- (nnc.south |- llm2.north);
            
            \draw [arrow] (scoren1.north |- llm2.south) -- (scoren1.north);
            \draw [arrow] (n2d.north |- llm2.south) -- (n2d.north);
            \draw [arrow] (nnd.north |- llm2.south) -- (nnd.north);
            
        \end{tikzpicture}
    }
    \caption{Research pipeline illustrating how process models (behavioral abstractions) are derived from sublogs containing only $k$ events. These models are explained by LLM$_1$ and scored by LLM$_2$ with the full model $M$ as a reference.}
    \label{fig:pipeline}
\end{figure}

%% file: additions/graph.tex
\begin{figure}[ht!]
    \centering
    \begin{tikzpicture}
        \begin{semilogxaxis}[
            width=0.80\textwidth,
            height=90,
            scale only axis,
            xlabel={Number of events},
            ylabel={Score},
            xmin=4, xmax=500000,
            ymin=4, ymax=9,
            xtick={10,100,1000,10000,100000},
            ytick={4,5,6,7,8,9,10},
            legend pos = south east,
            ymajorgrids=true, grid style=dashed,
        ]
            \addplot[red]
                coordinates {
                    (5, 4.60) (10, 4.84) (20, 5.93) (50, 6.77) (100, 6.87) (1000, 6.96)
                    (10000, 7.20) (100000, 7.81) (314160, 8.01)
                };
            \addplot[blue]
                coordinates{
                    (5, 4.90) (10, 4.60) (20, 6.13) (50, 5.80) (100, 6.13) (1000, 7.2)
                    (10000, 6.76) (100000, 7.93) (244384, 7.87)
                };
            \addplot[orange]
                coordinates{
                    (5, 4.60) (10, 5.00) (20, 5.80) (50, 5.93) (100, 6.67) (1000, 7.13)
                    (10000, 6.97) (100000, 7.93) (317072, 8.13)
                };
            \addplot[green]
                coordinates{
                    (5, 4.60) (10, 5.00) (20, 5.97) (50, 6.33) (100, 6.10) (1000, 6.66)
                    (10000, 6.87) (100000, 8.00) (345520, 7.88)
                };
            \addplot[purple]
                coordinates{
                    (5, 4.67) (10, 5.00) (20, 5.40) (50, 6.27) (100, 6.60) (1000, 6.40)
                    (10000, 7.03) (100000, 7.95) (296184, 7.87)
                };
            \legend{411, 412, 413, 421, 422}
    
            \draw[color=gray] (axis cs:5, 3.67) -- (axis cs:5, 5.33);
            \draw[color=gray] (axis cs:10, 4) -- (axis cs:10, 6.67);
            \draw[color=gray] (axis cs:20, 5) -- (axis cs:20, 7);
            \draw[color=gray] (axis cs:50, 5) -- (axis cs:50, 7.33);
            \draw[color=gray] (axis cs:100, 5) -- (axis cs:100, 7.67);
            \draw[color=gray] (axis cs:1000, 5.33) -- (axis cs:1000, 7.8);
            \draw[color=gray] (axis cs:10000, 6) -- (axis cs:10000, 8);
            \draw[color=gray] (axis cs:100000, 7.67) -- (axis cs:100000, 8.33);
        \end{semilogxaxis}
    \end{tikzpicture}
    \caption{Average LLM-assigned explanation scores per number of input events ($k$). Each line corresponds to one experiment from Table~\ref{tab:tracesandlogs}. Error bars represent score variability across five explanation evaluations per $k$.}
    \label{fig:results}
\end{figure}
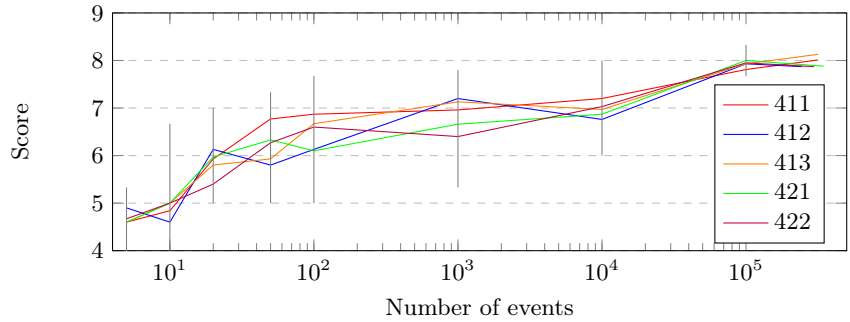

%% file: additions/tables.tex
\begin{table}[ht]
\centering
\resizebox{0.70\textwidth}{!}{%
\begin{tabular}{|c|ccccc|c|ccccc|c|ccccc|c|}
\hline
    Experiment & 
    \multicolumn{6}{c|}{5 logs} &
    \multicolumn{6}{c|}{10 logs} &
    \multicolumn{6}{c|}{20 logs} \\ 
&
    \multicolumn{6}{r|}{\textit{Avg}} &
    \multicolumn{6}{r|}{\textit{Avg}} &
    \multicolumn{6}{r|}{\textit{Avg}} \\ \hline
411 &
    5.0 & 4.3 & 4.3 & 5.0 & 4.3 & 4.6 &
    5.2 & 4.3 & 4.3 & 5.7 & 4.7 & 4.8 &
    6.3 & 6.0 & 5.3 & 5.7 & 6.3 & 5.9 \\
412 &
    5.3 & 5.2 & 4.7 & 4.7 & 4.7 & 4.9 &
    4.3 & 5.0 & 4.3 & 4.3 & 5.0 & 4.6 &
    6.0 & 7.0 & 6.0 & 6.0 & 5.7 & 6.1 \\
413 & 
    5.0 & 4.3 & 4.3 & 4.7 & 4.7 & 4.6 &
    5.7 & 4.3 & 4.3 & 4.7 & 6.0 & 5.0 &
    5.3 & 6.3 & 5.7 & 5.3 & 6.3 & 5.8 \\
421 &
    5.0 & 4.7 & 4.7 & 5.0 & 4.7 & 4.8 &
    5.7 & 4.7 & 4.7 & 5.0 & 4.7 & 5.0 &
    6.0 & 6.3 & 6.3 & 6.3 & 5.7 & 6.1 \\
422 &
    5.0 & 4.7 & 4.7 & 5.3 & 3.7 & 4.7 &
    5.7 & 5.7 & 5.0 & 4.7 & 4.0 & 5.0 &
    5.3 & 5.3 & 5.0 & 5.3 & 6.0 & 5.4 \\ \hline    
Experiment &    \multicolumn{6}{c|}{50 logs} &
    \multicolumn{6}{c|}{100 logs} &
    \multicolumn{6}{c|}{1,000 logs} \\
&
    \multicolumn{6}{r|}{\textit{Avg}} &
    \multicolumn{6}{r|}{\textit{Avg}} &
    \multicolumn{6}{r|}{\textit{Avg}} \\ \hline
411 &
    7.3 & 7.3 & 5.3 & 7.2 & 6.7 & 6.8 &
    7.7 & 7.0 & 7.7 & 6.7 & 5.3 & 6.9 &
    5.3 & 7.0 & 7.7 & 7.8 & 7.0 & 7.0 \\
412 &
    6.7 & 5.3 & 6.0 & 5.0 & 6.0 & 5.8 &
    5.0 & 7.0 & 5.3 & 6.7 & 6.7 & 6.1 &
    7.3 & 7.5 & 7.7 & 7.5 & 6.0 & 7.2 \\
413 & 
    6.0 & 6.0 & 6.3 & 6.0 & 5.3 & 5.9 &
    7.7 & 7.3 & 6.0 & 7.0 & 5.3 & 6.7 &
    6.7 & 7.0 & 7.0 & 7.3 & 7.7 & 7.1 \\
421 &
    4.0 & 4.7 & 4.3 & 5.3 & 4.7 & 4.6 &
    5.3 & 5.3 & 5.7 & 4.3 & 4.3 & 5.0 &
    7.3 & 5.7 & 6.7 & 7.0 & 6.7 & 6.7 \\
422 &
    6.7 & 5.3 & 6.7 & 7.3 & 5.3 & 6.3 &
    6.7 & 7.3 & 6.0 & 4.7 & 7.3 & 6.6 &
    6.0 & 6.7 & 6.0 & 7.3 & 6.0 & 6.4 \\ \hline
    Experiment & 
    \multicolumn{6}{c|}{10,000 logs} &
    \multicolumn{6}{c|}{100,000 logs} &
    \multicolumn{6}{c|}{Full log} \\
&
    \multicolumn{6}{r|}{\textit{Avg}} &
    \multicolumn{6}{r|}{\textit{Avg}} &
    \multicolumn{6}{r|}{\textit{Avg}} \\ \hline
411 &
    7.3 & 6.3 & 7.3 & 7.7 & 7.3 & 7.2 &
    8.0 & 7.7 & 8.0 & 7.7 & 7.7 & 7.8 &
    7.7 & 8.0 & 8.7 & 8.0 & 7.7 & 8.0 \\
412 &
    7.7 & 6.0 & 6.3 & 7.8 & 6.0 & 6.8 &
    8.0 & 8.0 & 7.7 & 8.0 & 7.7 & 7.9 &
    7.7 & 8.3 & 7.7 & 8.0 & 7.7 & 7.9 \\
413 & 
    7.3 & 7.2 & 7.0 & 7.3 & 6.0 & 7.1 &
    7.7 & 8.0 & 8.0 & 8.0 & 8.0 & 7.9 &
    8.0 & 8.3 & 8.3 & 8.0 & 8.0 & 8.1 \\
421 &
    7.3 & 6.7 & 6.7 & 6.0 & 7.7 & 6.9 &
    8.0 & 8.0 & 8.0 & 8.0 & 8.0 & 8.0 &
    7.7 & 8.0 & 7.7 & 8.0 & 8.0 & 7.9 \\
422 &
    6.0 & 7.3 & 8.0 & 6.0 & 7.8 & 7.0 &
    7.7 & 7.7 & 8.3 & 7.7 & 8.3 & 8.0 &
    7.7 & 7.7 & 8.0 & 8.0 & 8.0 & 7.9 \\
    \hline
\end{tabular}
}
\caption{Per-experiment explanation scores by input event count ($k$). Each row reports the five scores from LLM$_2$ for different values of $k$ (5 to full log), across five experiments (411–422). Each column group corresponds to a different event count; the final column in each group shows the average across the five runs. 
}
\label{tab:results1}
\end{table}

%% file: main.bbl
\begin{thebibliography}{10}
\providecommand{\url}[1]{\texttt{#1}}
\providecommand{\urlprefix}{URL }
\providecommand{\doi}[1]{https://doi.org/#1}

\bibitem{book-aalst-2007}
van~der Aalst, W.M.P., Reijers, H.A., Weijters, A.J.M.M., van Dongen, B.F., De~Medeiros, A.K.A., Song, M., Verbeek, H.M.W.: Business process mining: An industrial application. Information systems  \textbf{32},  713--732 (2007)

\bibitem{howmuchdataisenough}
Bauer, M., Senderovich, A., Gal, A., Grunske, L., Weidlich, M.: {How much event data is enough? A statistical framework for process discovery}. In: Advanced Information Systems Engineering. pp. 239--256 (2018)

\bibitem{bemthuis2021data}
Bemthuis, R., Koot, M., Mes, M.R.K., Bukhsh, F.A., Iacob, M.E., Meratnia, N.: {Data underlying the paper: An agent-based process mining architecture for emergent behavior analysis}. 4TU.Centre for Research Data  (2021). \doi{10.4121/12708839.V2}

\bibitem{berti2024pmllmbenchmarkevaluatinglargelanguage}
Berti, A., Kourani, H., van~der Aalst, W.M.P.: {PM-LLM-benchmark: Evaluating large language models on process mining tasks}. In: International Conference on Process Mining. pp. 610--623. Springer (2024)

\bibitem{berti2024evaluatingllminpm}
Berti, A., Kourani, H., Häfke, H., Li, C.Y., Schuster, D.: Evaluating large language models in process mining: Capabilities, benchmarks, and evaluation strategies, p. 13–21. Springer Nature Switzerland (2024). \doi{10.1007/978-3-031-61007-3_2}

\bibitem{berti2023leveraginglargelanguagemodels}
Berti, A., Qafari, M.S.: {Leveraging Large Language Models (LLMs) for process mining (Technical Report)} (2023), \url{https://arxiv.org/abs/2307.12701}

\bibitem{berti2019processminingpythonpm4py}
Berti, A., van Zelst, S.J., {W.M.P. van der Aalst}: {Process Mining for Python (PM4Py): Bridging the Gap Between Process- and Data Science} (2019), \url{https://arxiv.org/abs/1905.06169}

\bibitem{surveyonevaluationofllm}
Chang, Y., Wang, X., Wang, J., Wu, Y., Yang, L., Zhu, K., Chen, H., Yi, X., Wang, C., Wang, Y., Ye, W., Zhang, Y., Chang, Y., Yu, P.S., Yang, Q., Xie, X.: A survey on evaluation of large language models. ACM Transactions on Intelligent Systems and Technology  \textbf{15}(3) (2024). \doi{10.1145/3641289}

\bibitem{emamjome}
Emamjome, F., Andrews, R., ter Hofstede, A.H.M.: {A Case Study Lens on Process Mining in Practice}. In: On the Move to Meaningful Internet Systems: OTM 2019 Conferences. pp. 127--145. Cham (2019)

\bibitem{grattafiori2024llama3herdmodels}
Grattafiori, A., Dubey, A., Jauhri, A., et~al.: {The Llama 3 herd of models} (2024), \url{https://arxiv.org/abs/2407.21783}

\bibitem{grohs2023largelanguagemodelsaccomplish}
Grohs, M., Abb, L., Elsayed, N., Rehse, J.R.: Large language models can accomplish business process management tasks. In: International conference on business process management. pp. 453--465. Springer (2023)

\bibitem{jessen2023chitchatdeeptalkprompt}
Jessen, U., Sroka, M., Fahland, D.: {Chit-Chat or Deep Talk: Prompt Engineering for Process Mining} (2023), \url{https://arxiv.org/abs/2307.09909}

\bibitem{kubrak2024explanatory}
Kubrak, K., Botchorishvili, L., Milani, F., Nolte, A., Dumas, M.: Explanatory capabilities of large language models in prescriptive process monitoring. In: International Conference on Business Process Management. pp. 403--420. Springer (2024)

\bibitem{lashkevich2024llm}
Lashkevich, K., Milani, F., Avramenko, M., Dumas, M.: {LLM-assisted optimization of waiting time in business processes: A prompting method}. In: International Conference on Business Process Management. pp. 474--492. Springer (2024)

\bibitem{leemans2014process}
Leemans, S.J.J., Fahland, D., van~der Aalst, W.M.P.: Process and deviation exploration with inductive visual miner. In: 12th International Conference on Business Process Management, BPM 2014. pp. 46--50. CEUR-WS. org (2014)

\bibitem{eventlogpreprocessing}
Marin-Castro, H.M., Tello-Leal, E.: Event log preprocessing for process mining: A review. Applied Sciences  \textbf{11}(22) (2021). \doi{10.3390/app112210556}

\bibitem{rebmann2024evaluating}
Rebmann, A., Schmidt, F.D., Glava{\v{s}}, G., van Der~Aa, H.: {Evaluating the ability of LLMs to solve semantics-aware process mining tasks}. In: 2024 6th International Conference on Process Mining (ICPM). pp. 9--16. IEEE (2024)

\bibitem{viner2021processminingsoftwarecomparison}
Viner, D., Stierle, M., Matzner, M.: A process mining software comparison (2021), \url{https://arxiv.org/abs/2007.14038}

\bibitem{wang2024history}
Wang, Z., Chu, Z., Doan, T.V., Ni, S., Yang, M., Zhang, W.: {History, development, and principles of Large Language Models: An introductory survey}. AI and Ethics pp. 1--17 (2024)

\bibitem{pminthelarge}
{W.M.P. van der Aalst}: Process mining in the large, pp. 353--385. Springer Berlin Heidelberg, Berlin, Heidelberg (2016). \doi{10.1007/978-3-662-49851-4_12}

\bibitem{bpmindustrialapplication}
{W.M.P. van der Aalst}, Reijers, H., Weijters, A., {van Dongen}, B., {Alves de Medeiros}, A., Song, M., Verbeek, H.: Business process mining: An industrial application. Information Systems  \textbf{32}(5),  713--732 (2007). \doi{10.1016/j.is.2006.05.003}

\end{thebibliography}
